\documentclass[conference]{IEEEtran}
\IEEEoverridecommandlockouts
\usepackage{cite}
\usepackage{amsmath,amssymb,amsfonts}
\usepackage{algorithm}
\usepackage{algpseudocode} 
\usepackage{graphicx}
\usepackage{textcomp}
\usepackage{xcolor}
\usepackage{bm}
\usepackage{subcaption}

\def\BibTeX{{\rm B\kern-.05em{\sc i\kern-.025em b}\kern-.08em
    T\kern-.1667em\lower.7ex\hbox{E}\kern-.125emX}}
\begin{document}

\title{Learning Conditionally Independent Transformations using Normal Subgroups in Group Theory}

\author{\IEEEauthorblockN{Kayato Nishitsunoi\IEEEauthorrefmark{1},
Yoshiyuki Ohmura\IEEEauthorrefmark{1},
Takayuki Komatsu\IEEEauthorrefmark{1} and
Yasuo Kuniyoshi\IEEEauthorrefmark{1}\IEEEauthorrefmark{2}}
\IEEEauthorblockA{\IEEEauthorrefmark{1}School of Information Science and Technology, The University of Tokyo, Tokyo, Japan}
\IEEEauthorblockA{\IEEEauthorrefmark{2}Next Generation Artificial Intelligence Research Center (AI Center), The University of Tokyo, Tokyo, Japan}
\IEEEauthorblockA{Email: \{nishitsunoi, ohmura, komatsu, kuniyosh\}@isi.imi.i.u-tokyo.ac.jp}}

\maketitle

\begin{abstract}
Humans develop certain cognitive abilities that enable them to recognize objects and their transformations without explicit supervision. This suggests the importance of developing models for unsupervised representation learning. In unsupervised representation learning, a fundamental challenge is to separate different transformations in learned feature representations. Although algebraic approaches have been explored to achieve this separation, a comprehensive theoretical framework remains underdeveloped. To address this, researchers have proposed methods that decompose independent transformations based on algebraic independence, which generalizes the mathematical notion of independence. However, these methods primarily focus on transformations that commute, which means that the order in which two transformations are applied does not affect the final result. The methods do not naturally extend to cases where transformations are conditionally independent but noncommutative, where changing the order of application leads to different outcomes.  To extend current representation learning frameworks, we draw inspiration from Galois theory, where the decomposition of groups through normal subgroups provides an approach for the analysis of structured transformations. Normal subgroups naturally extend the concept of commutativity under certain conditions and offer a foundation for the categorization of transformations, even when they do not commute. In this paper, we propose a novel approach to representation learning that extends the theoretical framework of transformation decomposition by leveraging normal subgroups from Galois theory. Our method enables the separation of transformations that are conditionally independent, even in the absence of commutativity. In the experiment, we apply our method to the learning of geometric transformations in images and demonstrate that it successfully categorizes conditionally independent transformations, such as rotation and translation, in an unsupervised manner. This result suggests that the decomposition of groups via normal subgroups, as established in Galois theory, is closely related to the categorization of transformations in representation learning.
\end{abstract}

\begin{IEEEkeywords}
unsupervised learning,
representation learning theory, 
transformation decomposition, 
normal subgroup in group theory,
NeuralODE
\end{IEEEkeywords}

\section{Introduction}
Humans can recognize features and the motion of objects by disentangling complexly structured visual information. Such visual recognition develops without explicit instruction from others. The ability to learn object representations in an unsupervised manner is a core aspect of human intelligence and the acquisition of such a recognition system has been a long-standing goal in artificial intelligence research \cite{singh2021illiterate}.

In recent years, research on unsupervised representation learning has increasingly focused on using statistical independence to obtain representations that are disentangled along different axes \cite{kingma2013auto, higgins2017beta, zhu2021commutative}. These studies have made it possible to separate certain features, such as an object's color, size, and movement. However, challenges remain, such as over-separating features into too many axes or mixing different feature representations within a single axis, which makes clear feature categorization difficult.
Several methods for representation learning have been proposed, but their theoretical foundations are not yet fully established. By contrast, Higgins et al. introduced a group-theoretic perspective, which defines disentangled representations such that transformations within the learned vector space remain independent of each other \cite{higgins2018towards}. The acquisition of such invariant transformations is a fundamental challenge in pattern recognition \cite{otsu2011argus} and plays a crucial role in the explanation of the persistence of recognition \cite{pitts1947we}. Despite their importance, how to obtain these invariant transformations and connect them with practical representation learning methods, such as variational autoencoders \cite{kingma2013auto}, remains an unresolved problem.

To bridge this gap between theory and practical methodology, Ohmura et al. proposed an approach that leverages \textit{algebraic independence} \cite{simpson2018category}, which is a generalized mathematical concept of independence, to learn transformations invariant to specific vector representations \cite{ohmura2025unsupervised}. They assumed a model that, unlike conventional representation learning methods, which formulate reconstruction-based learning where the input and output are the same, explains changes between different sensory inputs using multiple transformations. The algebraic independence imposes a constraint that includes commutativity, which means that the order of the application of transformations does not affect the final outcome. This commutativity condition is deeply connected to the acquisition of independence, which ensures that transformations do not interfere with each other. Their method uses the algebraic independence constraints to separate independent transformations, such as color and shape, into distinct vector spaces. 
However, because this approach assumes that transformations are commutative, it does not apply to cases where transformations interact in a noncommutative manner, such as rotation and translation. For example, rotation and translation do not commute because the order in which they are applied affects the final result. If an object is first rotated around the origin and then translated, the final position differs from the case where translation is applied first, followed by rotation. This is because translation shifts the coordinate system, which effectively changes the center of rotation. 

This raises a broader problem: transformation categorization cannot be fully explained by independence alone. Existing independence-based approaches struggle to describe decomposition structures beyond simple factorization. Meanwhile, humans naturally distinguish between transformations such as rotation and translation, even though these do not satisfy algebraic independence. We can define such transformations that affect each other but remain distinguishable as conditionally independent transformations to capture the relationship. Although independence is a common principle in representation learning, it does not always align with human perception, which indicates the need for a more general framework.

In Galois theory, group decomposition provides an approach for the generalization of commutativity and classification of transformations in a structured manner. Galois introduced the concept of \textit{proper decomposition} \cite{singh1999last}, which later became formalized as a \textit{normal subgroup}, a subgroup that is invariant under conjugation by members of the group of which it is a part. Independence remains a useful principle in representation learning; however, it is insufficient to capture the full range of transformation structures. The extension of the theory of representation learning to incorporate \textit{structured decomposition} offers a promising direction for the separation of transformations beyond simple independence constraints.

The contribution of this study is to formulate representation learning using Galois' group decomposition theory and propose a novel representation learning method based on it. Furthermore, we experimentally showed that the proposed model can recognize conditionally independent transformations, such as rotation and translation, in an image sequence in an unsupervised manner. This contribution is expected to generalize the theory of representation learning, thereby providing a clearer perspective on transformation structures.

\section{Preliminaries}
In this section, we introduce fundamental concepts from group theory that are essential for our method, and focus on the distinction between commutativity and normal subgroups.

A group $G$ with a binary operation $\circ$ is said to be \textbf{commutative} (or \textbf{abelian}) if for all elements $a, b \in G$, the operation satisfies:
\begin{align}
a \circ b = b \circ a.
\end{align}
This property ensures that the order of application of elements does not affect the result of the operation. Commutativity is related to algebraic independence \cite{simpson2018category}.

By contrast, the concept of a \textbf{normal subgroup} generalizes commutativity. A subgroup $N \subset G$ is said to be \textbf{normal} in $G$ if, for any element $g \in G$, the left and right cosets of $N$ coincide:
\begin{align}
\label{normal}
gN = Ng.
\end{align}
Cosets play a fundamental role in understanding normal subgroups. Given a subgroup $N$ of a group $G$, the left cosets of $N$ in $G$ are defined as:
\begin{align}
gN = \{ g \circ n \mid n \in N \}, \quad \text{for } g \in G.
\end{align}
Similarly, the right cosets are given by:
\begin{align}
Ng = \{ n \circ g \mid n \in N \}, \quad \text{for } g \in G.
\end{align}
If $N$ is normal, the left and right cosets coincide, which ensures a well-defined partition of $G$ into disjoint subsets. This structured decomposition, which Galois referred to as \textbf{proper decomposition}, helps to analyze the group's internal structure and its behavior under transformations \cite{singh1999last}. Using a normal subgroup, a natural separation of the group can be achieved.

An intuitive interpretation of (\ref{normal}) appears in transformation systems. Consider an object undergoing both rotation and translation. In this case, rotation and translation are not commutative as discussed in the previous section. By contrast, if a change in the center of rotation is allowed, it is possible to satisfy (\ref{normal}) by setting \(g\) as translation and \(N\) as rotation. This can be interpreted as the rotational transformation represented by \(N\) in (\ref{normal}) being a transformation with a different center of rotation on the left-hand side and the right-hand side. Using a normal subgroup, it becomes possible to represent (\ref{normal}), which expresses a conditionally independent relationship.

To achieve this, it is essential to understand how normal subgroups can be derived. It is well known that normal subgroups can be obtained from group homomorphisms. Based on this, we now turn our attention to group homomorphisms. A function $f: G \to G'$ between groups $G$ and $G'$ with binary operations $\circ$ and $\cdot$, respectively, is called a \textbf{homomorphism} if it satisfies:
\begin{align}
f(g_1 \circ g_2) = f(g_1) \cdot f(g_2), \quad \forall g_1, g_2 \in G.
\end{align}
This condition ensures that the group structure is preserved under $f$. A key property of homomorphisms is the existence of their \textbf{kernel}, defined as:
\begin{align}
\ker(f) := \{ g \in G \mid f(g) = e \}
\end{align}
where $e$ is the identity element in $G'$. The kernel of a homomorphism consists of elements in $G$ that do not affect the structure of $G'$, which means that $\ker(f)$ represents transformations in $G$ that leave all elements in $G'$ invariant. It is known that \(ker(f)\) is a normal subgroup of the group \(G\). In this study, we aim to learn conditionally independent transformations by learning this \(ker(f)\).

These observations suggest an important extension in representation learning models. In conventional settings, commutativity constraints often enforce strict independence between features. However, by considering $\ker(f)$ and its normality, we can generalize these constraints beyond simple independence to conditional independence structures.
This relaxation permits structured dependencies within transformations, which make it possible to encode richer representations beyond naive independence assumptions.

\section{Formulation of transformation separation using homomorphism}\label{homo_method}
In this section, we propose an approach to unsupervisedly separating conditionally independent transformations, such as rotational and translational transformations, in a given sequence by leveraging homomorphisms to obtain normal subgroups.

Our study considers a model that learns these transformations from a sequence \(S\) that consists of \(T\) frame images \(\{\bm x_{0}, \bm{x_{1}}, \dots , \bm{x_{T-2}}, \bm{x_{T-1}}\}\). Suppose this sequence contains two types of transformations, denoted as \(g\) and \(v\). We define a group \(G\) whose elements are pairs of transformations \((g, v)\) and formulate a learning method within this group structure. In this group, if we define the transformation from \(\bm{x_{i}}\) to \(\bm{x_{j}}\) as \((g_{i,j}, v_{i,j})\), this \((g_{i,j}, v_{i,j})\) performs image transformation according to the following equation:
\begin{align}
    \label{group_operator}
    \bm{x_{j}} = (v_{i,j} \circ g_{i,j}) \bm{x_{i}}.
\end{align}
This equation implies that \(\bm{x_{j}}\) is obtained by first applying the transformation \(g_{i,j}\) to \(\bm{x_{i}}\), followed by the transformation \(v_{i,j}\).

Our model learns a function \(f\) such that \(v = f(g, v)\). Because we assume that \(f\) is a homomorphism, it satisfies the following equation:
\begin{align}
    &f((g_{i}, v_{i}) \circ (g_{j}, v_{j})) = f((g_{i}, v_{i})) \cdot f((g_{j}, v_{j})) .
\end{align}
If we  define \((g_{k}, v_{k}) = (g_{i}, v_{i}) \circ (g_{j}, v_{j})\), this implies
\begin{align}
    \label{homo_theory}
    v_{k} = v_{i} \cdot v_{j}.
\end{align}
In this case, the normal subgroup, which we can define using the kernel of \(f\), consists of transformations that satisfy \(f(g, v) = e\), which implies \(v = e\). Thus, the set of transformations \((g, e) = g\) forms a normal subgroup \(N\) of \(G\). If we let \(ker(f) = N\), the transformation \(v\) represents the residual transformation group in \(G\) that cannot be fully expressed by \(N\) alone.

To enforce this structure, we impose constraints on our proposed model to ensure that it satisfies the homomorphism property. This allows \(g\) to be learned as the transformations that form a normal subgroup, whereas \(v\) is learned as the residual transformations obtained by quotienting using the normal subgroup.

Furthermore, many types of geometric transformations form Lie groups, each consisting of differentiable transformations parameterized by a continuous variable \(\lambda\). If we express a Lie group transformation \(T\) using a scalar parameter \(\lambda\) as \(T(\lambda)\), the group satisfies the following properties \cite{otsu2011argus}:
\begin{align}
    &T(\mu)T(\lambda) = T(\mu + \lambda) = T(\lambda)T(\mu) \quad \text{(\(\lambda, \mu \in \mathbb{R}\))} \notag \\
    \label{lie_group}
    &T(0) = \textit{I} \qquad \text{(Identity operator)} \\
    &T(\lambda)^{-1} = T(-\lambda) \qquad \text{(Inverse operator)}. \notag
\end{align}
Given this property, we construct a model that learns the transformations \(g\) and \(v\) under the assumption that they form Lie groups.

\section{Learning transformations as Lie groups using NeuralODE}
In previous studies, researchers explored the unsupervised learning of geometric transformations under the assumption that they form Lie groups using NeuralODE \cite{takada2022disentangling, takatsuki2023unsupervised}.

In this section, we introduce the coordinate transformation approach proposed by Takada et al. as a method for learning geometric transformations using NeuralODE \cite{takada2022disentangling, takatsuki2023unsupervised}. In particular, we extend their method to enable the creation of conditional transformations in this study.

NeuralODE \cite{chen2018neural} is a neural network that solves ordinary differential equations (ODEs) using integration methods such as Euler's method and the Runge--Kutta method. It can be regarded as a continuous-depth network, where the traditionally discrete layers are replaced by an infinitesimally small depth interval. If the depth of the model is considered as a continuous parameter of the transformation group, NeuralODE naturally satisfies the properties of the Lie group mentioned in (\ref{lie_group}).

In this study, we design geometric image transformations as Lie group transformations using NeuralODE. Because geometric image transformations can be determined by understanding how each pixel moves, we implement pixel-wise coordinate transformations using NeuralODE. The ODE of the pixel-wise coordinate transformation is formulated as follows:

\begin{align}
    \label{4:ode}
    \frac{\mathrm{d}}{\mathrm{d}t}
    \begin{bmatrix}
    x\left( t \right) \\ y\left( t \right)
    \end{bmatrix}
    &=
    f\left(
    \begin{bmatrix}
    x\left( t \right) \\ y\left( t \right)
    \end{bmatrix}
    \right)  = A
    \begin{bmatrix}
    x\left( t \right) \\ y\left( t \right)
    \end{bmatrix}
     + \bm{b} + \bm{c}
\end{align}
where \(x\) and \(y\) denote the pixel coordinates, \(A \in \mathbb{R}^{2 \times 2}\), and \(\bm{b}, \bm{c} \in \mathbb{R}^{2 \times 1}\).

We assume that \(A\) and \(\bm{b}\) are shared across inter-frame transformations in a given dataset, whereas \(\bm{c}\) is a learnable variable specific to each sequence and inter-frame transformation. This design enables the model to generate rotational transformations with different centers of rotation and translational transformations with different orientations from a common ODE for each transformation.

Thus, we obtain the Lie group coordinate transformation in this study by integrating (\ref{4:ode}) over the integration time \(\lambda \in \mathbb{R}\), which is also a learnable parameter:
\begin{align}
    \label{4:integral}
    \begin{bmatrix}
    x^\lambda \\ y^\lambda
    \end{bmatrix}
    =
    F(\lambda) \left(
    \begin{bmatrix}
    x^0 \\ y^0
    \end{bmatrix}
    \right)
    &= 
    \int_{0}^{\lambda} f\left(\begin{bmatrix}
        x\left( t \right) \\ y\left( t \right)
        \end{bmatrix}\right) dt + \begin{bmatrix}
            x^0 \\ y^0
            \end{bmatrix}
\end{align} 
where \((x^0, y^0)\) denotes the initial coordinates before the transformation and \((x^\lambda, y^\lambda)\) denotes the transformed coordinates computed by NeuralODE.

\section{Transformation learning model}
In this section, we describe the details of the model that learns to separate transformations from image sequences using the homomorphism-based transformation separation and NeuralODE-based coordinate transformation methods introduced in the previous sections. Specifically, we construct a model that learns two types of transformations \(g\) and \(v\), which are present in the image sequences. In this study, the transformations from the \(i\)-th frame to the \(j\)-th frame are represented as \(g(\lambda^g_{i,j}, \bm{c}^g_{i,j})\) and \(v(\lambda^v_{i,j}, \bm{c}^v_{i,j})\). This notation indicates that the transformations \(g\) and \(v\) are computed using NeuralODE, where each transformation is determined by the integration time \(\lambda_{i,j}\) and conditional variable \(\bm{c}_{i, j}\) between the \(i\)-th and \(j\)-th image frames based on their respective ODEs.

\subsection{Architecture of the pixel-wise transformer}
In this section, we describe a method for performing pixel-wise image transformation based on the coordinate transformation using NeuralODE, as introduced in the previous section.

In this study, we implement image transformation using the image sampling technique introduced in spatial transformer networks \cite{jaderberg2015spatial}, which enables gradient-computable geometric transformations. The image sampling process is expressed as follows using the coordinate notation \((x^0, y^0)\) and \((x^\lambda, y^\lambda)\) introduced in the previous section:
\begin{align}
    \label{2:STN_sampling}
    V^{c}_{i} = \sum^{H, W}_{n, m} U^{c}_{nm} max(0, 1-|x^{\lambda}_i-m|) max(0, 1-|y^{\lambda}_i-n|)
\end{align}
where \(U^{c}_{nm}\) represents the value at location \((n, m)\) in channel \(c\) of the input image and \(V^{c}_{i}\) is the output value for pixel \(i\) of \((x^0, y^0)\) coordinates in channel \(c\).
This equation indicates that the image transformation is performed using linear interpolation based on the sampling grid \((x^{\lambda}, y^{\lambda})\), which specifies where the feature (e.g., RGB values) of each pixel of the transformed image (\((x^0, y^0)\) coordinates) is mapped from the original image.

By defining the image transformation in this manner, we can compute the gradient with respect to the sampling grid as follows: 
\begin{align}
    \label{2:STN_differential2}
    \frac{\partial V^{c}_{i}}{\partial x^{\lambda}_{i}} = \sum^{H,W}_{n,m} U^{c}_{nm} max(0, 1-|y^{\lambda}_i-n|)  
    \begin{cases}
        0 &\text{if $ |m-x^{\lambda}_{i}| \geq 1 $} \\
        1 &\text{if $ m \geq x^{\lambda}_{i} $} \\
        -1 &\text{if $ m < x^{\lambda}_{i} $}.
      \end{cases}
\end{align}
A similar equation holds for \(\frac{\partial V^{c}_{i}}{\partial y^{\lambda}_{i}}\).

\subsection{Encoder-based prediction of transformation parameters}
\begin{figure*}[h]
  \centering
  \begin{subfigure}[b]{0.4\linewidth}
    \centering
    \includegraphics[width=7.5cm,clip]{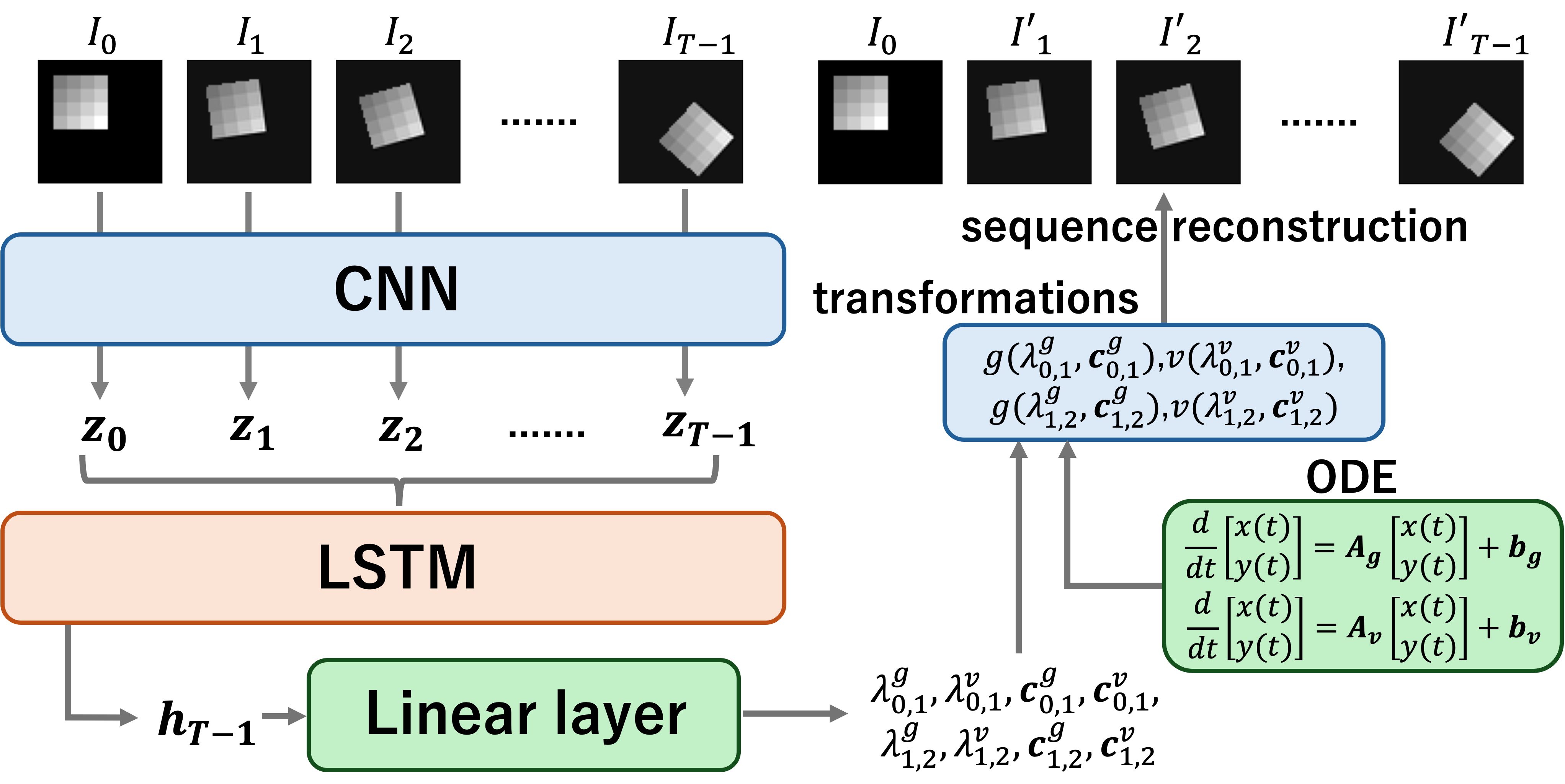}
    \caption{}
    \label{fig:encoder_flow}
  \end{subfigure}
  \hfill
  \begin{subfigure}[b]{0.5\linewidth}
    \centering
    \includegraphics[width=9cm,clip]{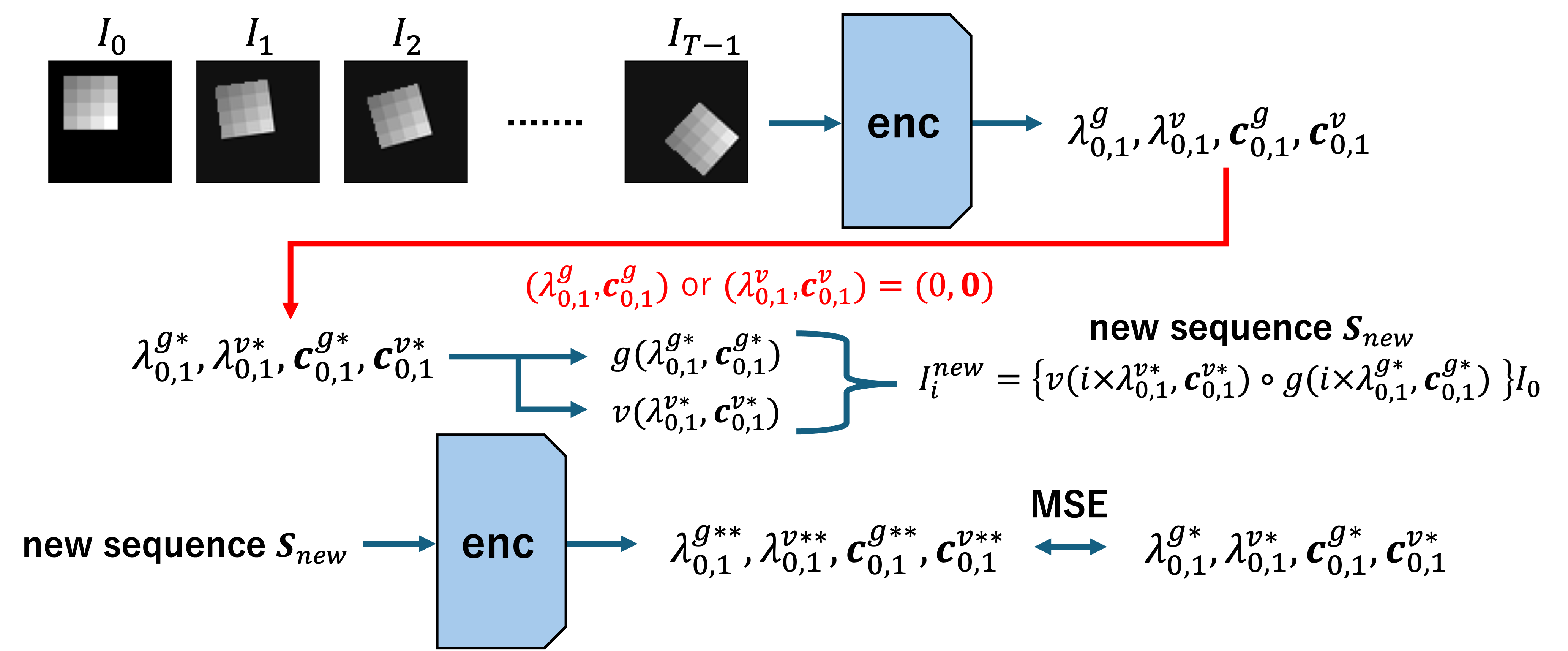}
    \caption{}
    \label{fig:ssl_flow}
  \end{subfigure}
  \caption{Overview of the proposed model. (a) Architecture of the encoder and reconstruction learning. CNNs output feature vector \(\bm{z}_{i}\) for frame \(I_{i}\) and transformation parameters \(\lambda, \bm{c}\) are calculated using LSTM and a linear layer. Then we reconstruct the input sequence by creating geometric transformations \(g\) and \(v\) from the transformation parameters and the two ODEs. (b) Architecture of self-supervised learning. We set the elements of the computed transformation parameters \(\lambda\) and \(\bm{c}\) to 0, a new sequence is generated where either \(g\) or \(v\) becomes an identity transformation, and from this sequence, the encoder learns to predict the transformation parameters this model has created.}
\end{figure*}

To learn transformations, it is necessary to estimate not only the ODE parameters but also the transformation magnitudes \(\lambda\) and the conditional variables \(\bm{c}\) for each transformation. Therefore, we construct an encoder \(f_{\theta}\) based on a combination of convolutional neural networks (CNNs) and long short-term memory (LSTM) networks \cite{hochreiter1997long} to compute a sequence feature vector \(h_{T-1}\) and a linear layer to compute the transformation parameters from \(h_{T-1}\). The process is illustrated in Fig. \ref{fig:encoder_flow}. The parameters \(\theta\) of the encoder are learned jointly with the remainder of the model.

\subsection{Reconstruction of the input sequence}
In this study, we assume that each sequence contains two types of geometric transformations and use image reconstruction learning to infer these transformations. Let \(I_{t}\) denote the \(t\)-th frame in the sequence. Referring to (\ref{group_operator}), the reconstruction of frame \(I_{t}\) can be expressed as follows:
\begin{align}
    \label{4:sequence_reconst}
    I^{\prime}_{i} = \{v(i\times\lambda^v_{0,1}, \bm{c}^v_{0,1}) \circ g(i\times\lambda^g_{0,1}, \bm{c}^g_{0,1})\}I_{0}.
\end{align}
This equation implies that if the transformation in a given sequence is stationary, the sequence can be reconstructed by the application of the transformation between Frame 0 and Frame 1, with the integration time scaled by a factor of \(i\).

Therefore, the sequence reconstruction loss is as follows:
\begin{align}
    \label{4:seq_reconst_loss}
    \mathcal{L}_{recon} = \sum_{i=1}^{T-1} MSE(I_i, I^{\prime}_{i})
\end{align} 
where $MSE$ represents the mean square error.

\subsection{Homomorphism constraint}
As introduced in Section \ref{homo_method}, we aim to construct a model that can theoretically partition the transformation by leveraging homomorphism. To achieve this, it is necessary to impose constraints on the learned transformations to satisfy (\ref{homo_theory}). Specifically, in this study, this equation applies to transformations between the first three consecutive frames in a given sequence. Let \(v_{k}\), \(v_{i}\), and \(v_{j}\) in (\ref{homo_theory}) represent the transformations from the zeroth frame to the second frame, from the first frame to the second frame, and from the zeroth frame to the first frame, respectively. Then the following equation holds:
\begin{align}
    \label{4:homo_image_trans1}
    &v_kI_{0} = v(2\times\lambda^v_{0,1}, \bm{c}^v_{0,1})I_{0} \\
    \label{4:homo_image_trans2}
    &\{v_i\cdot v_j\}I_0 = \{v(\lambda^v_{1,2}, \bm{c}^v_{1,2}) \cdot v(\lambda^v_{0,1}, \bm{c}^v_{0,1})\}I_0.
\end{align} 
Note that, in this study, \(\cdot\) represents an operator, which indicates that it applies image transformations in the same manner as \(\circ\).
Consequently, the homomorphism constraint can be expressed as follows based on (\ref{homo_theory}):
\begin{align}
    \label{4:homo_relation}
    v_kI_{0} = \{v_{i} \cdot v_j\}I_0.
\end{align} 
Thus, the loss function for the homomorphism constraint in this study is defined as follows:
\begin{align}
    \label{homo_loss}
    \mathcal{L}_{homo} = MSE(v_kI_0, \{v_{i} \cdot v_j\}I_0).
\end{align} 
Additionally, because \(v(\lambda^v_{1,2}, \bm{c}^v_{1,2})\) must be able to reconstruct the second frame from the first frame, the following frame reconstruction loss is required:
\begin{align}
    &I''_{2} = \{v(\lambda^v_{1,2}, \bm{c}^v_{1,2}) \circ g(\lambda^g_{1,2}, \bm{c}^g_{1,2})\}I_{1} \\
    \label{4:seq_reconst_loss2}
    &\mathcal{L}_{recon2} = MSE(I_2, I''_{2}).
\end{align} 
The parameters \(\lambda^{g}_{1,2}\), \(\bm{c}^{g}_{1,2}\), \(\lambda^{v}_{1,2}\), \(\bm{c}^{v}_{1,2}\) are computed from a common encoder along with the other transformation parameters \(\lambda^{g}_{0,1}\), \(\bm{c}^{g}_{0,1}\), \(\lambda^{v}_{0,1}\), \(\bm{c}^{v}_{0,1}\).

\subsection{Self-supervised learning for invariance learning}
In this section, we describe self-supervised learning for learning homomorphism \(f\) and kernel \(ker(f)\). 

From the definition of homomorphism in Section \ref{homo_method}, \(f(g,v)=v\) must hold for any \(g\). To ensure this, we create a sequence in which \(g\) varies while \(v\) remains invariant during training. Specifically, we generate a sequence where \(g\) is set to \(e\) and perform self-supervised learning to predict the transformation parameters of the altered transformation from the sequence. If \(g\) and \(v\) were to learn unintended representations, for example, both representing rotation, this loss would increase.

Similarly, for learning \(ker(f)\)  (\(=g\)), we ensure that \(g\) does not contain any components of \(v\). To achieve this, we explicitly enforce \(f(g, v)=v=e\), which means that the transformation that corresponds to \(g\) is that obtained when \(v\) is set as the identity transformation. This learning process is analogous to self-supervised learning described earlier, where \(g\) was set as the identity, but now applied to the case where \(v\) is the identity.

The flow of this self-supervised learning process is illustrated in Fig. \ref{fig:ssl_flow}. First, for a given input sequence, we obtain the predicted transformation parameters \(\lambda^g_{0,1}\), \(\bm{c}^g_{0,1}\), \(\lambda^v_{0,1}\), \(\bm{c}^v_{0,1}\) through the encoder. We then set only the components corresponding to either \(g\) or \(v\) to zero, thereby obtaining modified transformation parameters \(\lambda^{g*}_{0,1}\), \(\bm{c}^{g*}_{0,1}\), \(\lambda^{v*}_{0,1}\), \(\bm{c}^{v*}_{0,1}\). Using (\ref{new_seq}), we generate a new sequence \(S_{new}\) (\(S_{new} := \{I_0, I^{new}_1, \cdots,I^{new}_{T-1}\}\)), in which one of the transformations is set to the identity transformation: 
\begin{align}
    \label{new_seq}
    &I^{new}_i = \{v(i\times\lambda^{v*}_{0,1}, \bm{c}^{v*}_{0,1}) \circ g(i\times\lambda^{g*}_{0,1}, \bm{c}^{g*}_{0,1})\}I_{0}. 
\end{align}
By re-predicting the transformation parameters from \(S_{new}\) using the encoder, we achieve self-supervised learning. Let \(\lambda^{g**}_{0,1}\), \(\bm{c}^{g**}_{0,1}\), \(\lambda^{v**}_{0,1}\), \(\bm{c}^{v**}_{0,1}\) be the transformation parameters predicted by the encoder from \(S_{new}\). Then the constraint for self-supervised learning is defined as follows:
\begin{align}
    \label{ssl_const}
    \begin{bmatrix}
        \lambda^{g*}_{0,1}, \bm{c}^{g*}_{0,1}, \lambda^{v*}_{0,1}, \bm{c}^{v*}_{0,1}
    \end{bmatrix}
    =
    \begin{bmatrix}
        \lambda^{g**}_{0,1}, \bm{c}^{g**}_{0,1}, \lambda^{v**}_{0,1}, \bm{c}^{v**}_{0,1}
    \end{bmatrix}. 
\end{align}
Consequently, the loss for self-supervised learning can be formulated as follows:
\begin{align}
    \label{ssl_loss}
    \mathcal{L}_{ssl}=\sum MSE(\begin{bmatrix}
        \lambda^{g*}_{0,1}, \bm{c}^{g*}_{0,1}, \lambda^{v*}_{0,1}, \bm{c}^{v*}_{0,1}
    \end{bmatrix}, \notag \\
    \begin{bmatrix}
        \lambda^{g**}_{0,1}, \bm{c}^{g**}_{0,1}, \lambda^{v**}_{0,1}, \bm{c}^{v**}_{0,1}
    \end{bmatrix}). 
\end{align}
At each training step, we perform self-supervised learning for both cases: one in which only \(g\) is set to \(e\) and another in which only \(v\) is set to \(e\). Thus, the sigma notation in (\ref{ssl_loss}) represents the sum of both self-supervised learning losses.

\subsection{Isometric transformation constraint}
Additionally, because we assume an isometric transformation as the geometric transformation, the relative distance between pixels of the coordinate that is the input to the NeuralODE and the relative distance between pixels of the transformed coordinate must be equal. This constraint is expressed as follows:
\begin{align}
    \label{4:distance_pixel_1}
    &D^0_{ij,kl} = \|(x^{0}_{i,j}, y^{0}_{i,j}) - (x^{0}_{k,l}, y^{0}_{k,l}) \|_{2} \\
    \label{4:distance_pixel_2}
    &D^\lambda_{ij,kl} = \|(x^{\lambda}_{i,j}, y^{\lambda}_{i,j}) - (x^{\lambda}_{k,l}, y^{\lambda}_{k,l}) \|_{2} \\
    \label{4:loss_structure_invariance}
    &\mathcal{L}_{trans} = \sum MSE(D^{0}, D^{\lambda}) \quad \text{\((D^{0}, D^{\lambda} \in \mathbb{R}^{HW \times HW})\)}
\end{align} 
where \(D^{0}\) and \(D^{\lambda}\) are distance matrices that represent the pixel-wise distances of the input and output coordinates of the NeuralODE, respectively. Their dimensions are \(HW \times HW\), where \(H\) and \(W\) denote the height and width of the image in pixels. Each element at position \((ij, kl)\) in the matrices \(D^{0}\) and \(D^{\lambda}\) represents the distance between the coordinates of the pixel at \((i, j)\) and that of the pixel at \((k, l)\) in each grid. In (\ref{4:loss_structure_invariance}), the sigma notation is used to sum the mean squared error of the distance matrices for all transformations \(g\) and \(v\) generated by the model. Additionally, because the coordinates \((x^0,y^0)\) represent the standard grid coordinates before being transformed by NeuralODE, \(D^0\) remains the same matrix for all transformations.

\subsection{\(\bm{c}\) norm constraint}
Additionally, as an auxiliary loss for learning the ODE parameters \((A, \bm{b})\), we incorporate the squared norm of \(\bm{c}\), as shown in (\ref{4:loss_c_norm}) for all \(\bm{c}\) values calculated by the model. This loss serves to reduce the expressive capacity of \(\bm{c}\), which encourages the differential equation before the mapping to represent the sequence transformation as much as possible. Furthermore, we introduce it to enhance the uniqueness of the solution because different combinations of \(\lambda\) and \(\bm{c}\) could otherwise result in the same translational transformation:

\begin{align}
    \label{4:loss_c_norm}
    \mathcal{L}_{\bm{c}} = \|\bm{c}^g_{0,1}\|_2 + \|\bm{c}^v_{0,1}\|_2 + \|\bm{c}^g_{1,2}\|_2+\|\bm{c}^v_{1,2}\|_2.
\end{align}

\subsection{Overall loss}
To summarize, the loss function for the transformer is as follows:

\begin{align}
    \label{4:loss_total}
    \mathcal{L}_{seq} = \alpha \mathcal{L}_{recon} + \beta \mathcal{L}_{recon2} + \gamma \mathcal{L}_{homo} + \delta \mathcal{L}_{ssl} \notag \\+ \epsilon \mathcal{L}_{trans} + \zeta \mathcal{L}_{\bm{c}}
\end{align} 
where \(\alpha, \beta, \gamma, \delta, \epsilon, \zeta \in \mathbb{R}\) represent the weights of each loss.

\section{Experiments}
\subsection{Dataset and experimental settings}
In the experiments, we aimed to verify whether transformation separation could be achieved solely based on motion, that is, regardless of the shape of the object in the sequence. Given this aim, we created two types of sequences and conducted transformation learning experiments for each sequence dataset.

Sequence 1 is shown in Fig. 2 as \textit{Given Sequence}. It consists of seven grayscale images of size \(64 \times 64\) pixels (1 channel), where a textured square of size \(30 \times 30\) pixels serves as the object and undergoes rotational and translational motion. We generated the object's texture by dividing the square into a \(4 \times 4\) grid (16 blocks) and assigning each block a different grayscale intensity. We determined the motion of the object by selecting a rotational angle per frame from \(\{8^{\circ},10^{\circ},12^{\circ}\}\) and translational shifts in both the \(x\) and \(y\) directions from \(\{-3,0,3\}\), with three values randomly chosen for each sequence. 

Sequence 2 is shown in Fig. 3 as \textit{Given Sequence}. It also consists of seven grayscale images of size \(64 \times 64\) pixels (1 channel), but in this case, the object is a textured semicircle with a radius of 8 pixels, which undergoes both rotational and translational motion. Each sequence contains two semicircles arranged to move together as a single object. We generated motion combinations in the same manner as those in Sequence 1.

In the experiments, the encoder consisted of four CNN layers whose kernel size was 3, stride was 2, and padding was 1. An LSTM with one hidden layer and one fully connected linear layer followed. We used the ReLU function for activation in the CNN layers. The channels of the convolution layers were 16, 32, 64, and 128. The size of the hidden layer of the LSTM was 128, and the input size and output size of the linear layer were 128 and 12 (the sum of the size of \(\lambda\) and \(\bm{c}\)), respectively.

For the integration computation in NeuralODE, we adopted the simplest method: the Euler method. To improve integration accuracy, we computed the integration interval \(\varDelta t\) as shown in 
\begin{align}
    \label{5:num_steps}
    \varDelta t &= \frac{\lambda}{\lfloor |\lambda| \times K \rfloor + 1}
\end{align}
where \(\lfloor \cdot \rfloor\) represents the floor function and the constant \(K\) allows the number of subdivisions of the integration interval to be adjusted according to the integration time. Using this approach, we improved integration accuracy while ensuring that the gradient of the integration time parameter \(\lambda\) could still be computed. We specifically set the constant \(K\) to 10 in our experiments.

In the experiments, we first performed initialization training for the encoder parameters \(\theta\), followed by sequence reconstruction learning and self-supervised learning. During the initialization of the encoder, we trained the model so that both \(g\) and \(v\) were recognized as identity transformations for all given sequences. Specifically, we set the initial values of the ODE parameters \(A\) and \(\bm{b}\) to zero. Accordingly, we initialized the encoder parameters such that its output became \(\lambda=1\) and \(\bm{c}=0\). (The reason that we set \(\lambda=1\) is that NeuralODE implemented in PyTorch cannot perform computations when \(\lambda=0\).) The loss function used in this training is defined as follows:
\begin{align}
    \mathcal{L}_{init} = MSE([\lambda^g_{0,1}, \bm{c}^g_{0,1}, \lambda^v_{0,1}, \bm{c}^v_{0,1}],[1, \bm{0}, 1, \bm{0}])\notag \\ + MSE([\lambda^g_{1,2}, \bm{c}^g_{1,2}, \lambda^v_{1,2}, \bm{c}^v_{1,2}],[1, \bm{0}, 1, \bm{0}]).
\end{align}
For this initialization phase, we used RAdam as the optimizer and set the learning rate for the encoder parameters to 0.0001. We set the number of training steps to 1,000 to ensure that the loss converged sufficiently.

For the reconstruction and self-supervised learning, we also used RAdam as the optimizer, with a learning rate of 0.0001 for both the ODE parameters \(A, \bm{b}\) and the encoder parameters \(\theta\). We initialized the elements of \(A\) and \(\bm{b}\) to zero before training. We set the weights for each term in the loss function to \(\alpha=1\), \(\beta=1\), \(\gamma=1\), \(\delta=0.1\), \(\epsilon=1\), \(\zeta=0.1\). We set the number of training steps to 20,000 to ensure that the overall loss converged sufficiently.

\subsection{Results}
\begin{table}[htbp]
\caption{ODE parameters for each transformer learned from Sequence 1.}
\begin{center}
\begin{tabular}[b]{@{}c@{}}
        \hline
        \textbf{Transformer \(g\)} \\
        \( A = \begin{bmatrix}
        2.80 \times 10^{-3} & -1.49 \times 10^{-1} \\
        1.48 \times 10^{-1} & 2.10 \times 10^{-3}
        \end{bmatrix} \)
        \hfill
        \( \bm{b} = \begin{bmatrix}
        -1.14 \times 10^{-2} \\
        -2.96 \times 10^{-2}
        \end{bmatrix} \) \\
        \\
        \textbf{Transformer \(v\)} \\
        \( A = \begin{bmatrix}
        3.59 \times 10^{-5} & 4.46 \times 10^{-4} \\
        3.60 \times 10^{-4} & 3.32 \times 10^{-4}
        \end{bmatrix} \)
        \hfill
        \( \bm{b} = \begin{bmatrix}
        4.22 \times 10^{-2} \\
        1.81 \times 10^{-2}
        \end{bmatrix} \) \\
        \hline
        \end{tabular}
\label{tab1}
\end{center}
\end{table}
\begin{table}[htbp]
\caption{ODE parameters for each transformer learned from Sequence 2.}
\begin{center}
\begin{tabular}[b]{@{}c@{}}
        \hline
        \textbf{Transformer \(g\)} \\
        \( A = \begin{bmatrix}
        7.08 \times 10^{-5} & -1.18 \times 10^{-1} \\
        1.16 \times 10^{-1} & -8.34 \times 10^{-4}
        \end{bmatrix} \)
        \hfill
        \( \bm{b} = \begin{bmatrix}
        1.11 \times 10^{-2} \\
        3.46 \times 10^{-2}
        \end{bmatrix} \) \\
        \\
        \textbf{Transformer \(v\)} \\
        \( A = \begin{bmatrix}
        1.10 \times 10^{-3} & -9.00 \times 10^{-4} \\
        2.40 \times 10^{-3} & 9.00 \times 10^{-4}
        \end{bmatrix} \)
        \hfill
        \( \bm{b} = \begin{bmatrix}
        -3.04 \times 10^{-2} \\
        1.11 \times 10^{-2}
        \end{bmatrix} \) \\
        \hline
        \end{tabular}
\label{tab2}
\end{center}
\end{table}

\begin{figure*}[htbp]
    \label{result1}
    \begin{tabular}{cc}
      \begin{minipage}[t]{0.4\linewidth}
        \centering
        \includegraphics[width=4cm,clip]{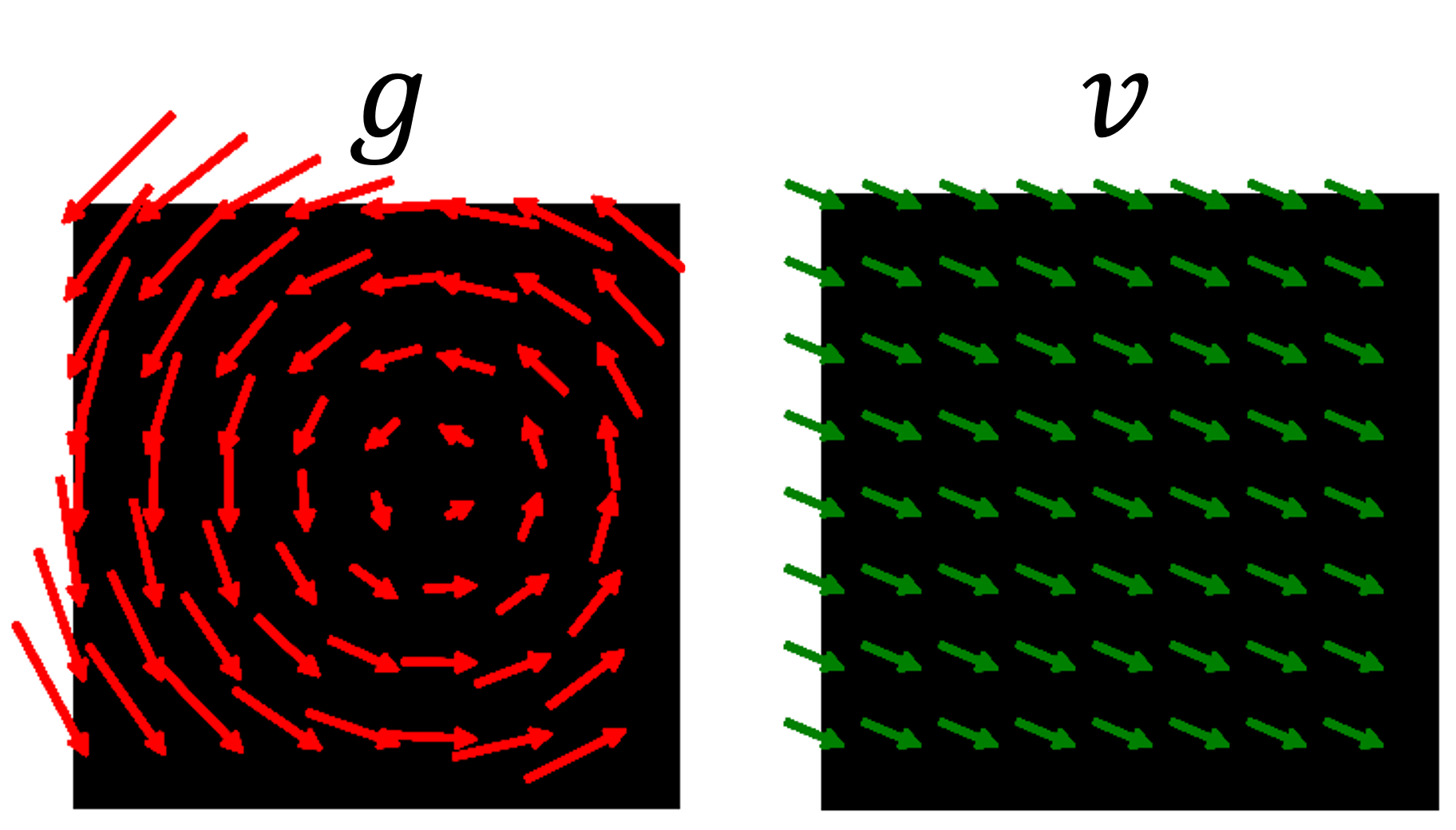}
        \subcaption{}
        \label{composite}
      \end{minipage} &
      \begin{minipage}[t]{0.45\linewidth}
        \centering
        \includegraphics[width=7cm,clip]{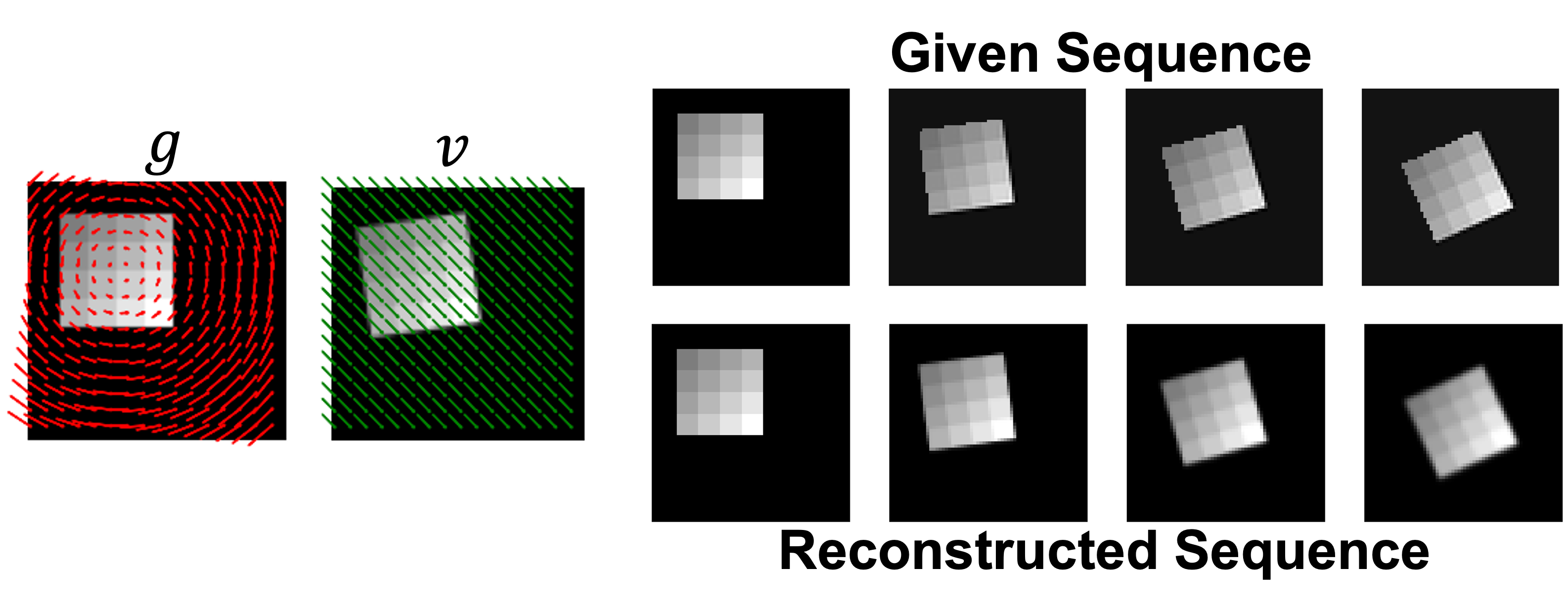}
        \subcaption{}
        \label{Gradation}
      \end{minipage} \\
   
      \begin{minipage}[t]{0.45\linewidth}
        \centering
        \includegraphics[width=7cm,clip]{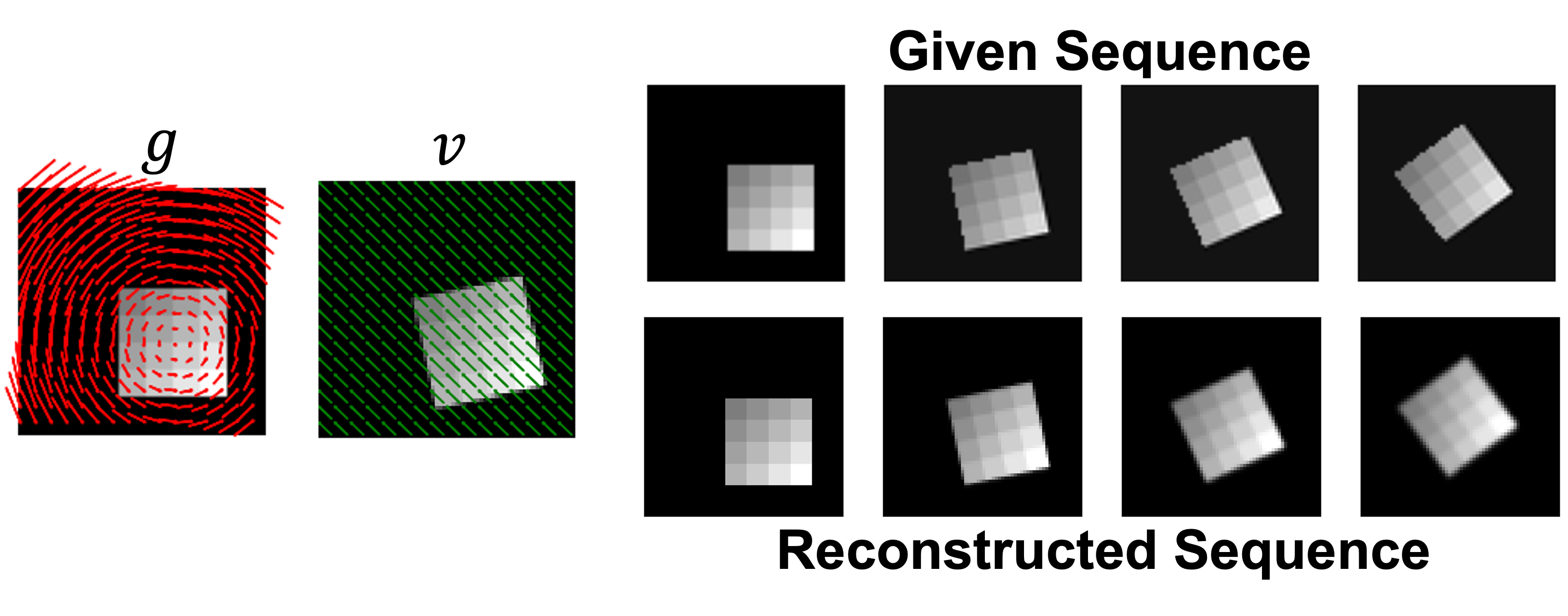}
        \subcaption{}
        \label{fill}
      \end{minipage} &
      \begin{minipage}[t]{0.45\linewidth}
        \centering
        \includegraphics[width=7cm,clip]{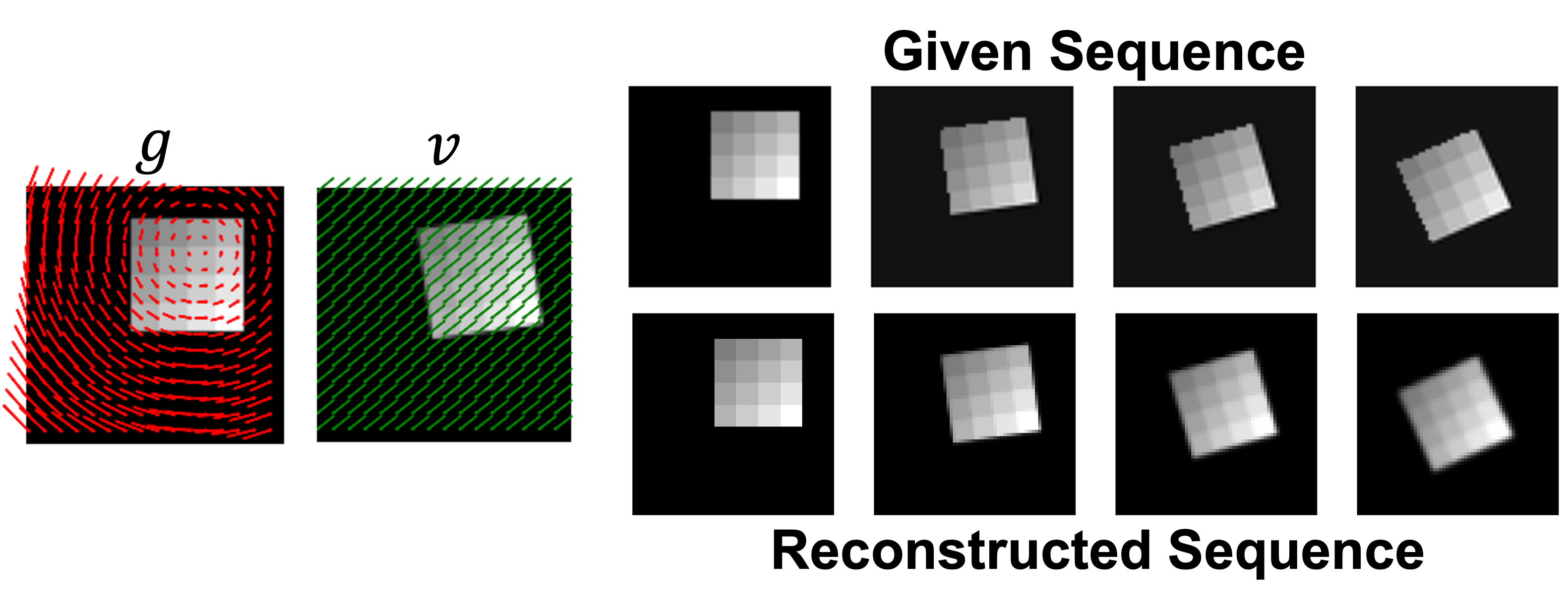}
        \subcaption{}
        \label{transform}
      \end{minipage} 
    \end{tabular}
     \caption{Two transformation fields learned from the Sequence 1 dataset. (a) Transformation fields computed by setting \((\lambda, \bm{c})=(2, \bm{0})\) for \(g\) and \((\lambda, \bm{c})=(-3, \bm{0})\) for \(v\), based on the learned ODE parameters for each transformation. (b), (c), (d) Transformations learned for each given sequence and the reconstructed sequences generated by the application of these transformations. \(g\) and \(v\) show the \(g(\lambda^g_{0,1}, \bm{c}^g_{0,1})\) and \(v(\lambda^v_{0,1}, \bm{c}^v_{0,1})\)  transformations, respectively, and the given sequence and reconstructed sequence each display only the first four frames out of the seven frames.}
  \end{figure*}

\begin{figure*}[htbp]
    \label{result2}
    \begin{tabular}{cc}
      \begin{minipage}[t]{0.4\linewidth}
        \centering
        \includegraphics[width=4cm,clip]{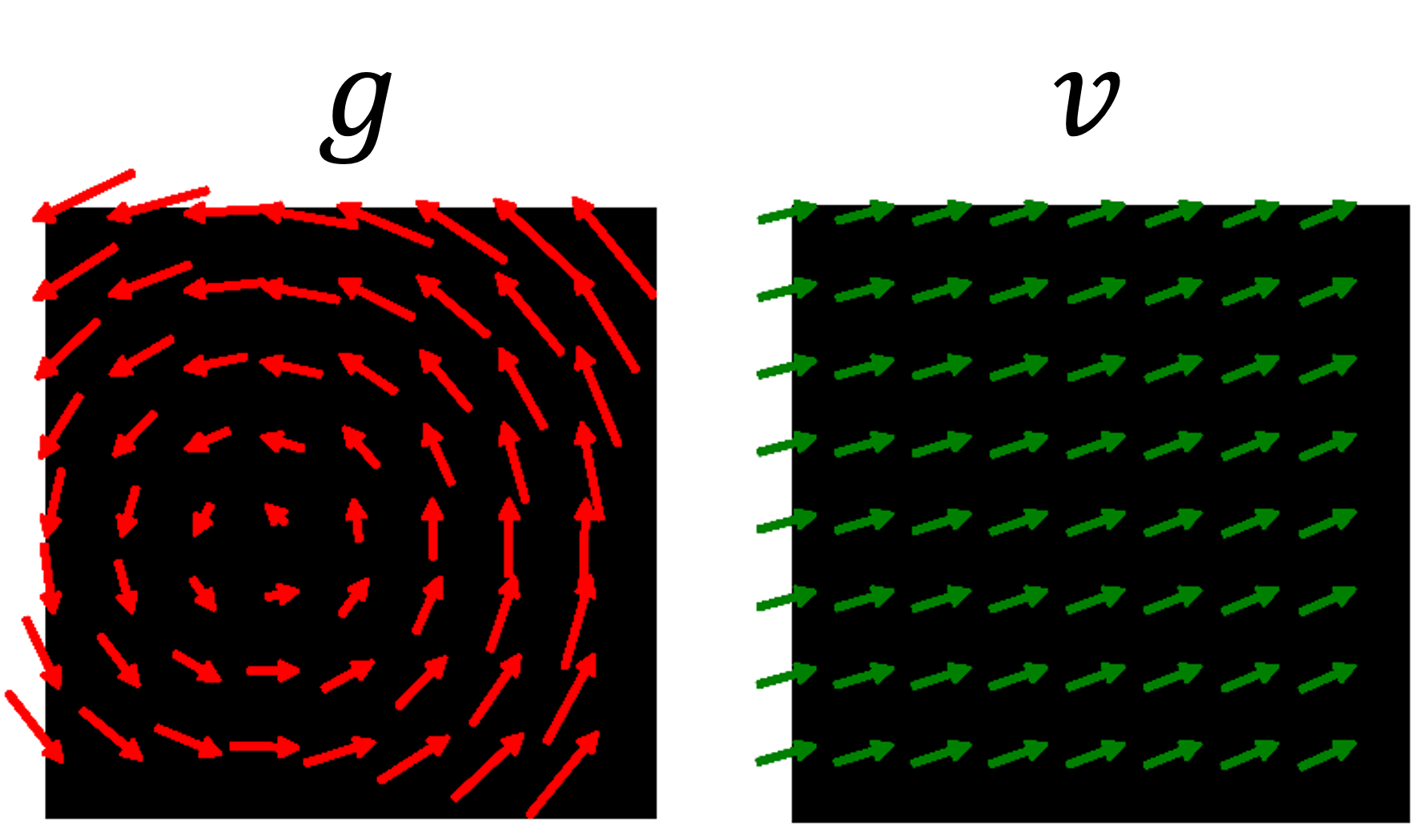}
        \subcaption{}
        \label{composite2}
      \end{minipage} &
      \begin{minipage}[t]{0.45\linewidth}
        \centering
        \includegraphics[width=7cm,clip]{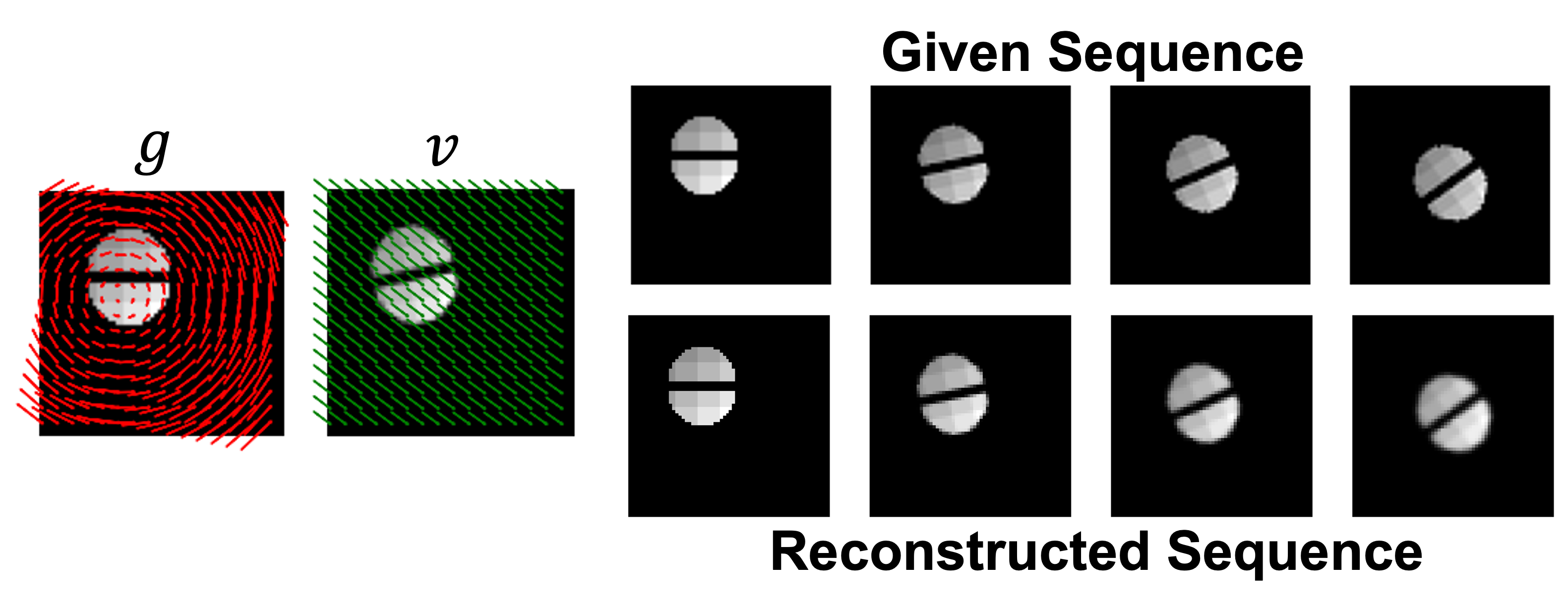}
        \subcaption{}
        \label{Gradation2}
      \end{minipage} \\
   
      \begin{minipage}[t]{0.45\linewidth}
        \centering
        \includegraphics[width=7cm,clip]{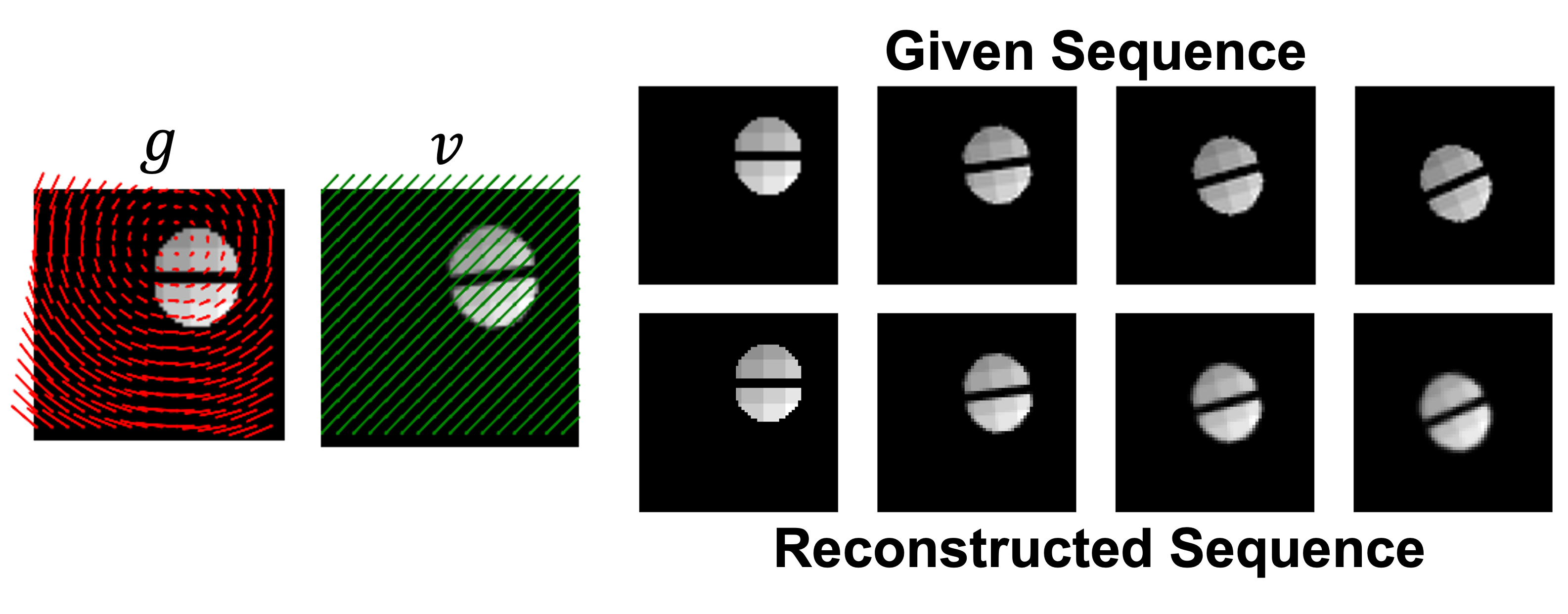}
        \subcaption{}
        \label{fill2}
      \end{minipage} &
      \begin{minipage}[t]{0.45\linewidth}
        \centering
        \includegraphics[width=7cm,clip]{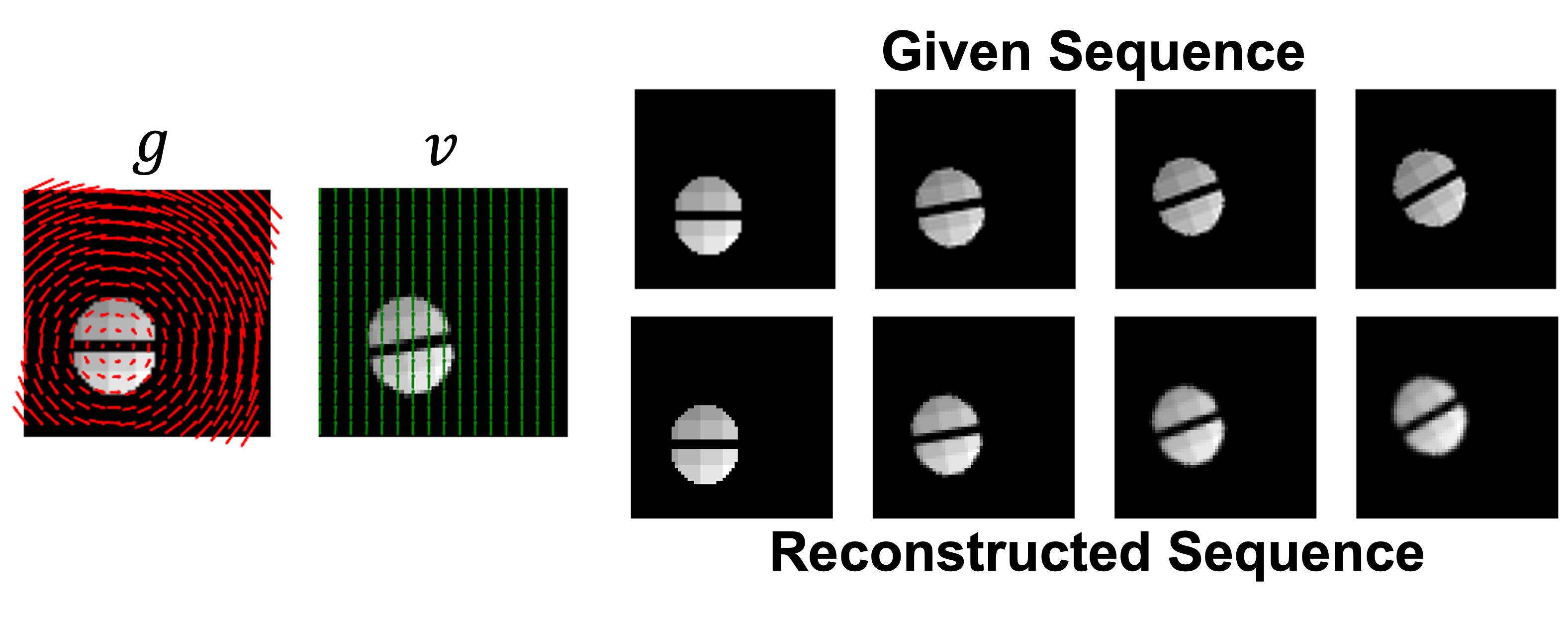}
        \subcaption{}
        \label{transform2}
      \end{minipage} 
    \end{tabular}
     \caption{Two transformation fields learned from the Sequence 2 dataset. (a) Transformation fields computed by setting \((\lambda, \bm{c})=(2, \bm{0})\) for \(g\) and \((\lambda, \bm{c})=(4, \bm{0})\) for \(v\), based on the learned ODE parameters for each transformation. (b), (c), (d) Transformations learned for each given sequence and the reconstructed sequences generated by the application of these transformations. \(g\) and \(v\) show the \(g(\lambda^g_{0,1}, \bm{c}^g_{0,1})\) and \(v(\lambda^v_{0,1}, \bm{c}^v_{0,1})\) transformations, respectively, and the given sequence and reconstructed sequence each display only the first four frames out of the seven frames.}
  \end{figure*}
Tables \ref{tab1} and \ref{tab2} show the learned ODE parameters \(A, \bm{b}\), and Fig. \ref{composite} and Fig. \ref{composite2} show the transformation flows obtained from the ODEs. Additionally, Fig. \ref{Gradation}, Fig. \ref{fill}, and Fig. \ref{transform}, and Fig. \ref{Gradation2}, Fig. \ref{fill2}, and Fig. \ref{transform2} show the learned transformation flows and the reconstructed images for each sequence, respectively.

From both tables, the differential equation learned as the first coordinate transformation module \(g\) exhibits a significant difference in magnitude between the (0,1) and (1,0) elements of \(A\) and its other elements. Additionally, the (0,1) element of \(A\) is similar to the multiplication of the (1,0) element of \(A\) by \(-1\), which indicates that this transformation represents a rotational transformation. This observation is further confirmed by Fig. \ref{composite} and Fig. \ref{composite2}.

Similarly, if we focus on the differential equation learned as the second transformation module \(v\), both tables show that the elements of \(A\) are minute compared with the value of the elements in \(\bm{b}\). This suggests that the influence of the translational component \(\bm{b}\) is dominant, and the transformation represents a translation. This is further corroborated by Fig. \ref{composite} and Fig. \ref{composite2}.

Furthermore, if we examine how the transformation fields \(g\) and \(v\) for each sequence were learned, we observe from Fig. \ref{Gradation}, Fig. \ref{fill}, and Fig. \ref{transform}, and Fig. \ref{Gradation2}, Fig. \ref{fill2}, and Fig. \ref{transform2} that the appropriate rotational and translational transformation fields were generated for each transformation.

These experimental results demonstrate that the proposed model successfully separated and learned two types of transformations, that is, rotation and translation, without prior knowledge of the transformations. Furthermore, because both rotation and translation were consistently recognized across two different sequence datasets, it is evident that the transformations were identified based on motion rather than object-specific features within the sequences.

This finding suggests that, with the application of the group decomposition approach based on normal subgroups to representation learning, it becomes possible to learn transformations that have conditional independence, which is an aspect that has been challenging in conventional representation learning.

\section{Conclusion}
In this study, we proposed a representation learning model that adapts the concept of normal subgroups, which is a fundamental concept in Galois' group decomposition theory, to representation learning methods. We based the proposed model on a coordinate transformation model using NeuralODE and structured it using sequence reconstruction, homomorphism constraints for learning normal subgroups, and self-supervised learning to aid their acquisition. The experimental results demonstrated that the model successfully separated and learned two conditionally independent transformations: rotation and translation. This indicates that the approach of group decomposition via normal subgroups has the potential to extend the conventional representation learning framework, which has traditionally been based on strict independence, to a framework capable of learning conditional independence.

However, in this study, we were limited to proposing a representation learning method that incorporates the concepts of homomorphism and normal subgroups to separate rotation and translation. We did not conduct a detailed investigation into how effective these mathematical elements are in learning conditional independence; hence, this remains a future research direction.

Additionally, in this study, we created conditional transformations by simply adding a two-dimensional vector \(\bm{c}\) to the ODE. However, it is also necessary to propose a more generalized approach that conditions the transformation on the matrix \(A\), which allows for a broader range of transformations.

Furthermore, although we designed our model based on the assumption of the existence of groups \(G\) and \(G'\), the binary operations that define these groups must be closed under the group structure. Because imposing such constraints is difficult, we did not address this issue in this study, leaving it as a topic for future work.

\section*{Acknowledgment}

This work was supported in part by the Grant-in-Aid for Scientific Research (A), under Grant JP22H00528. We thank Edanz (https://jp.edanz.com/ac) for editing a draft of this manuscript.

\bibliographystyle{IEEEtran}
\bibliography{main}

\begin{thebibliography}{10}
\providecommand{\url}[1]{#1}
\csname url@samestyle\endcsname
\providecommand{\newblock}{\relax}
\providecommand{\bibinfo}[2]{#2}
\providecommand{\BIBentrySTDinterwordspacing}{\spaceskip=0pt\relax}
\providecommand{\BIBentryALTinterwordstretchfactor}{4}
\providecommand{\BIBentryALTinterwordspacing}{\spaceskip=\fontdimen2\font plus
\BIBentryALTinterwordstretchfactor\fontdimen3\font minus \fontdimen4\font\relax}
\providecommand{\BIBforeignlanguage}[2]{{%
\expandafter\ifx\csname l@#1\endcsname\relax
\typeout{** WARNING: IEEEtran.bst: No hyphenation pattern has been}%
\typeout{** loaded for the language `#1'. Using the pattern for}%
\typeout{** the default language instead.}%
\else
\language=\csname l@#1\endcsname
\fi
#2}}
\providecommand{\BIBdecl}{\relax}
\BIBdecl

\bibitem{singh2021illiterate}
G.~Singh, F.~Deng, and S.~Ahn, ``Illiterate dall-e learns to compose,'' \emph{arXiv preprint arXiv:2110.11405}, 2021.

\bibitem{kingma2013auto}
D.~P. Kingma, ``Auto-encoding variational bayes,'' \emph{arXiv preprint arXiv:1312.6114}, 2013.

\bibitem{higgins2017beta}
I.~Higgins, L.~Matthey, A.~Pal, C.~P. Burgess, X.~Glorot, M.~M. Botvinick, S.~Mohamed, and A.~Lerchner, ``beta-vae: Learning basic visual concepts with a constrained variational framework.'' \emph{ICLR (Poster)}, vol.~3, 2017.

\bibitem{zhu2021commutative}
X.~Zhu, C.~Xu, and D.~Tao, ``Commutative lie group vae for disentanglement learning,'' in \emph{International Conference on Machine Learning}.\hskip 1em plus 0.5em minus 0.4em\relax PMLR, 2021, pp. 12\,924--12\,934.

\bibitem{higgins2018towards}
I.~Higgins, D.~Amos, D.~Pfau, S.~Racaniere, L.~Matthey, D.~Rezende, and A.~Lerchner, ``Towards a definition of disentangled representations,'' \emph{arXiv preprint arXiv:1812.02230}, 2018.

\bibitem{otsu2011argus}
N.~Otsu, ``Argus: Adaptive recognition for general use system-its theoretical construction and applications,'' \emph{Synthesiology English edition}, vol.~4, no.~2, pp. 75--86, 2011.

\bibitem{pitts1947we}
W.~Pitts and W.~S. McCulloch, ``How we know universals the perception of auditory and visual forms,'' \emph{The Bulletin of mathematical biophysics}, vol.~9, pp. 127--147, 1947.

\bibitem{simpson2018category}
A.~Simpson, ``Category-theoretic structure for independence and conditional independence,'' \emph{Electronic Notes in Theoretical Computer Science}, vol. 336, pp. 281--297, 2018.

\bibitem{ohmura2025unsupervised}
Y.~Ohmura, W.~Shimaya, and Y.~Kuniyoshi, ``Unsupervised categorization of similarity measures,'' \emph{arXiv preprint arXiv:2502.08098}, 2025.

\bibitem{singh1999last}
A.~Singh, ``The last mathematical testament of galois,'' \emph{Resonance}, pp. 93--100, 1999.

\bibitem{takada2022disentangling}
T.~Takada, W.~Shimaya, Y.~Ohmura, and Y.~Kuniyoshi, ``Disentangling patterns and transformations from one sequence of images with shape-invariant lie group transformer,'' in \emph{2022 IEEE International Conference on Development and Learning (ICDL)}.\hskip 1em plus 0.5em minus 0.4em\relax IEEE, 2022, pp. 54--59.

\bibitem{takatsuki2023unsupervised}
R.~Takatsuki, Y.~Ohmura, and Y.~Kuniyoshi, ``Unsupervised judgment of properties based on transformation recognition,'' in \emph{2023 IEEE International Conference on Development and Learning (ICDL)}.\hskip 1em plus 0.5em minus 0.4em\relax IEEE, 2023, pp. 409--414.

\bibitem{chen2018neural}
R.~T. Chen, Y.~Rubanova, J.~Bettencourt, and D.~K. Duvenaud, ``Neural ordinary differential equations,'' \emph{Advances in neural information processing systems}, vol.~31, 2018.

\bibitem{jaderberg2015spatial}
M.~Jaderberg, K.~Simonyan, A.~Zisserman \emph{et~al.}, ``Spatial transformer networks,'' \emph{Advances in neural information processing systems}, vol.~28, 2015.

\bibitem{hochreiter1997long}
S.~Hochreiter, ``Long short-term memory,'' \emph{Neural Computation MIT-Press}, 1997.

\end{thebibliography}

\end{document}